\documentclass[lettersize,journal]{IEEEtran}
\usepackage{amsmath,amsfonts}
\usepackage{algorithmic}
\usepackage{algorithm}
\usepackage{array}
\usepackage[caption=false,font=normalsize,labelfont=sf,textfont=sf]{subfig}
\usepackage{textcomp}
\usepackage{stfloats}
\usepackage{url}
\usepackage{verbatim}
\usepackage{graphicx}
\usepackage{cite}
\usepackage[square, numbers, sort&compress]{natbib} % For citation commands like \citep
\usepackage{multirow}
\usepackage{booktabs} % For better tables if needed later
\usepackage{hyperref}
\hypersetup{
    colorlinks=true,            % 激活彩色链接，去掉方框
    linkcolor=blue,             % 目录、公式链接颜色
    filecolor=magenta,          % 文件链接颜色
    urlcolor=cyan,              % 网址链接颜色
    citecolor=blue,             % 【关键】参考文献引用颜色，设置为蓝色
}
\begin{document}

\title{Generative Regression for Left Ventricular Ejection Fraction Estimation from Echocardiography Video}

\author{Jinrong Lv, Xun Gong \IEEEmembership{Member, IEEE}, Zhaohuan Li, and Weili Jiang
\thanks{This work is partially supported by National Natural Science Foundation of China (62376231), Fundamental Research Funds for the Central Universities (2682025ZTPY052, 2682023ZDPY001).  \textit{(Corresponding authors: Xun Gong)}}
\thanks{Jinrong Lv, Xun Gong, Weili Jiang are with the School of Computing and Artificial Intelligence, Southwest Jiaotong University, and the Manufacturing Industry Chain Collaboration Industrial Software Key Laboratory of Sichuan Province, Chengdu, Sichuan 610031, China (e-mail: lvjinrong@my.swjtu.edu.cn, lvjinrong.23@gmail.com; xgong@swjtu.edu.cn; jiangweili@swjtu.edu.cn, weilijiang111@gmail.com, machong@swjtu.edu.cn).}
\thanks{ZhaoHuan Li are with the School of Medicine, University of Electronic Science and Technology, Chengdu, Sichuan 610031, China (e-mail: lzhjx2007@uestc.edu.cn).}
}

\markboth{Journal of \LaTeX\ Class Files,~Vol.~14, No.~8, August~2021}%
{Shell \MakeLowercase{\textit{et al.}}: A Sample Article Using IEEEtran.cls for IEEE Journals}

\IEEEpubid{0000--0000/00\$00.00~\copyright~2021 IEEE}

\maketitle

\begin{abstract}
Estimating Left Ventricular Ejection Fraction (LVEF) from echocardiograms constitutes an ill-posed inverse problem. Inherent noise, artifacts, and limited viewing angles introduce ambiguity, where a single video sequence may map not to a unique ground truth, but rather to a distribution of plausible physiological values. Prevailing deep learning approaches typically formulate this task as a standard regression problem that minimizes the Mean Squared Error (MSE). However, this paradigm compels the model to learn the conditional expectation, which may yield misleading predictions when the underlying posterior distribution is multimodal or heavy-tailed—a common phenomenon in pathological scenarios.
In this paper, we investigate the paradigm shift from deterministic regression toward generative regression. We propose the Multimodal Conditional Score-based Diffusion model for Regression (MCSDR), a probabilistic framework designed to model the continuous posterior distribution of LVEF conditioned on echocardiogram videos and patient demographic attribute priors. Extensive experiments conducted on the EchoNet-Dynamic, EchoNet-Pediatric, and CAMUS datasets demonstrate that MCSDR achieves state-of-the-art performance. Notably, qualitative analysis reveals that the generation trajectories of our model exhibit distinct behaviors in cases characterized by high noise or significant physiological variability, thereby offering a novel layer of interpretability for AI-aided diagnosis. 
Code will be available at \url{https://github.com/lvmarch/MCSDR.git}.
\end{abstract}


% -----------------------------------------------------------------------------
\section{Introduction}
\label{sec:intro}

% \begin{figure}[t]
% \centering
% \includegraphics[width=0.95\columnwidth]{figs/intro.pdf} 
% \caption{\textbf{Illustration of the proposed LVEF probabilistic inference process.} The model learns to reverse the diffusion process (blue), transforming random noise (orange) into a valid LVEF sample. Crucially, this trajectory is guided by a fusion of echocardiogram videos and patient demographics (e.g., age, sex). By sampling multiple trajectories, we construct a full posterior distribution (green). A narrow distribution indicates high confidence supported by clear evidence, while a wide distribution captures the aleatoric uncertainty arising from image artifacts or ambiguity.}
% \label{fig_intro}
% \end{figure}

\begin{figure}[t]
\centering
\includegraphics[width=\columnwidth]{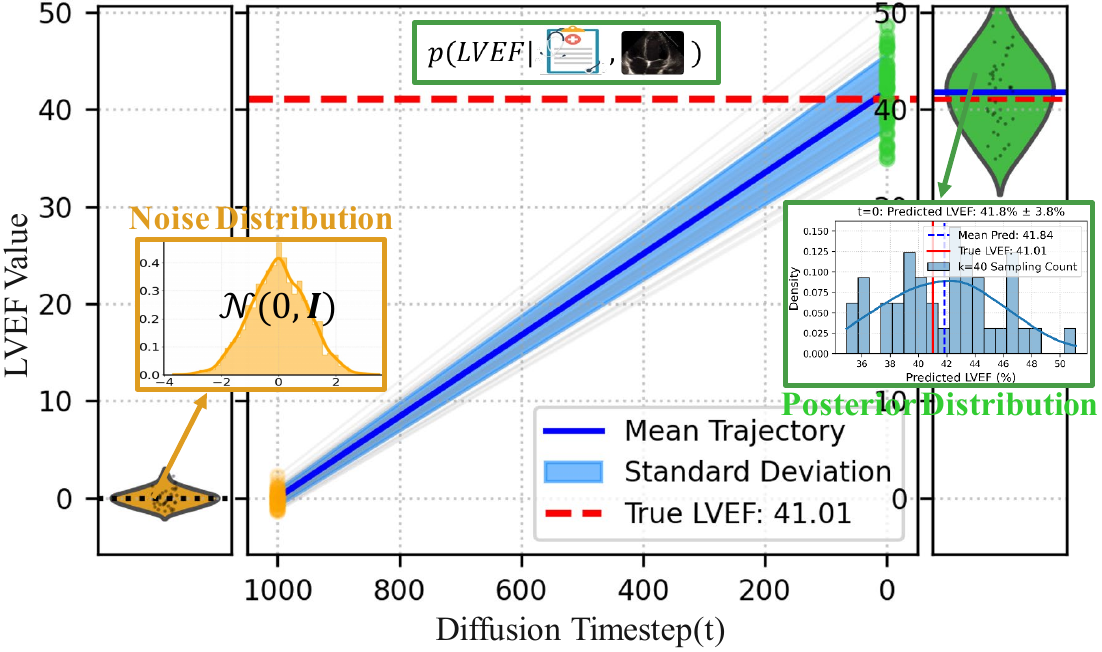} 
\caption{\textbf{Illustration of the proposed LVEF probabilistic inference process.} The model reverses the diffusion process (blue), transforming random noise (orange) into plausible LVEF samples. Crucially, this trajectory is guided by the fusion of echocardiogram videos and patient demographic attributes (e.g., age, sex). By sampling multiple trajectories, we approximate the full posterior distribution (green). A narrow distribution indicates high confidence supported by unambiguous evidence, and vice versa.}
\label{fig_intro}
\end{figure}

Left Ventricular Ejection Fraction (LVEF) is a standard metric in cardiology for diagnosing heart failure and guiding therapeutic decisions~\citep{circulation_1971_lvef, jacc_2003_lvef, jacc_2009_lvef, circulation_2022_lvef}. The automation of LVEF assessment using deep learning has seen steady progress, with recent models achieving performance comparable to human experts on large-scale datasets~\citep{miccai_2021_UVT, Miccai_2022_echognn, tmi_2023_semilvef, miccai_coreecho, tim_2025_fhfa}. However, the majority of existing approaches frame LVEF estimation as a standard deterministic regression problem, mapping an input video directly to a scalar value.

% We posit that this deterministic formulation may be mathematically suboptimal for echocardiography analysis. Medical imaging typically presents an \textit{ill-posed inverse problem}: due to inherent speckle noise, acoustic shadows, and variable acquisition angles, a single echocardiogram sequence may plausibly correspond to a range of LVEF values rather than a unique ground truth. Standard regression models trained with Mean Squared Error (MSE) or Mean Absolute Error (MAE) loss theoretically converge to the expectation of the conditional distribution $E[y|x]$. Consequently, when the data distribution is multimodal or heavy-tailed—common in pathological cases—deterministic models may output over-smoothed predictions that do not fully capture the underlying complexity or ambiguity of the diagnosis~\citep{bishop2006pattern}. Additionally, such models provide a point estimate without an intrinsic indication of reliability, a limitation in clinical settings where understanding prediction confidence is valuable.

We posit that this deterministic formulation is mathematically suboptimal for echocardiography analysis due to the \textit{ill-posed nature of the inverse problem}. Ultrasound imaging is inherently plagued by speckle noise, acoustic artifacts, and variability in operator-dependent views. Consequently, a single observed video sequence $\mathbf{x}$ may not map to a unique ground truth $y$, but rather to a distribution of plausible physiological values. Standard regression models trained with Mean Squared Error (MSE) or Mean Absolute Error (MAE) loss theoretically converge to the conditional expectation $\mathbb{E}[y|\mathbf{x}]$~\citep{bishop2006pattern}. While the mean is a statistically safe estimator, it can be misleading in medical contexts where the posterior distribution is multimodal or asymmetric. For instance, in cases of ambiguous wall motion, the conditional mean might yield an intermediate value that does not correspond to any valid pathological state, effectively blurring the diagnosis.
% \IEEEpubidadjcol

To address these challenges, we propose reframing LVEF estimation as a \textbf{Generative Regression} task. Instead of learning a direct mapping $f: \mathcal{X} \rightarrow \mathcal{Y}$, we aim to model the conditional posterior distribution $p(y | \mathcal{X}, \mathcal{A})$, where $\mathcal{X}$ represents the visual data and $\mathcal{A}$ denotes clinical priors. This probabilistic perspective allows for capturing the range of plausible outcomes given the evidence.

A second consideration regarding current \textit{Feature Extractor + Regression Head} architectures is the integration of non-visual priors. In clinical practice, a diagnosis is often informed by more than images alone; patient attributes such as age, sex, and weight provide Bayesian priors that can help constrain the scope of diagnostic inference ~\citep{EchoClip}. However, existing methodologies often overlook these readily available clinical contextual data.

In this paper, we introduce the \textbf{Multimodal Conditional Score-based Diffusion model for Regression (MCSDR)}. Leveraging the theoretical framework of Stochastic Differential Equations (SDEs)~\citep{ho2020denoising, DDIM2021, song2021scorebased, icml_2025_tutorial}, MCSDR learns to generate LVEF predictions by reversing a diffusion process. Starting from Gaussian noise, the model iteratively denoises the target variable, guided by a learned score function (gradient of the log-density).
To effectively parameterize this score function, we specifically architect the \textbf{Multimodal Conditional Score Network (MCSN)}. By integrating high-dimensional spatiotemporal visual representations with low-dimensional patient demographic attributes, the MCSN enables clinical priors to modulate the underlying gradient field, thereby implicitly steering the stochastic trajectory toward the ground-truth manifold.
 
This generative formulation offers two potential advantages. First, we aim to enhance prediction accuracy. By approximating the full distribution rather than a single mean, the diffusion process effectively accommodates the non-Gaussian distributions prevalent in medical data. Second, this framework provides an intrinsic mechanism for reliability assessment. By sampling multiple trajectories from the learned posterior distribution, we can visualize the dispersion of the solution space. A narrow distribution indicates a well-posed solution supported by unambiguous evidence, whereas a dispersed distribution reflects high aleatoric uncertainty, suggesting that the prediction warrants expert review.

Our main contributions are summarized as follows:
\begin{itemize}
    \item We introduce a diffusion-based framework for LVEF estimation, treating the task as a probabilistic inverse problem. This approach addresses the mean-regression tendency of deterministic models, allowing for the capture of complex posterior distributions.
    \item We design a conditioning mechanism that fuses high-dimensional echocardiogram features with tabular patient demographic attributes. This allows patient-specific priors to guide the solution of the inverse problem.
    \item We conduct comprehensive experiments on three public datasets: EchoNet-Dynamic, EchoNet-Pediatric, and CAMUS. MCSDR achieves state-of-the-art accuracy, particularly in challenging scenarios, and provides an interpretable analysis of prediction reliability.
\end{itemize}

%-----------------------------------------------------------------------------
\section{Related Work}
\label{sec:related_work}

% \subsection{Automated LVEF Estimation}
% The accurate quantification of Left Ventricular Ejection Fraction (LVEF) is pivotal for cardiovascular diagnosis~\citep{circulation_2022_lvef}. Early automated approaches relied on segmentation-based pipelines, calculating volumes from segmented masks~\citep{diag_2023_Batool}. 
% With the advent of deep learning, direct video-based regression has become the mainstream paradigm. The seminal work EchoNet-Dynamic~\citep{EchoNet-Dynamic} demonstrated expert-level performance using 3D CNNs, an approach subsequently extended to pediatric domains~\citep{EchoNet-Pediatric}. To capture complex spatiotemporal dependencies, recent works have incorporated advanced architectures, including Transformers~\citep{miccai_2021_UVT, miccai_2022_echocotr, miccai_coreecho} and Graph Neural Networks (GNNs)~\citep{Miccai_2022_echognn, Miccai_2022_echograph, tim_2025_fhfa}.

% However, despite these architectural innovations, the underlying learning paradigm remains largely \textbf{deterministic}. These methods typically model the conditional expectation $\mathbb{E}[y|\mathbf{x}]$, treating LVEF estimation as a point-mapping task. This formulation implicitly assumes a unimodal mapping and overlooks the aleatoric uncertainty inherent in noisy ultrasound imagery. Our work departs from this convention by reframing the task as a probabilistic inverse problem, utilizing a generative approach to capture the full distribution of plausible outcomes.

\subsection{Echocardiography Video Analysis and LVEF Estimation}
\label{sec:related_lvef}

Automated echocardiography video analysis has evolved from low-level signal processing to high-level semantic interpretation. Early methodologies focused on myocardial motion tracking using monogenic signals~\citep{tip2013}, optical flow~\citep{tip2005}, or sparse dictionary learning~\citep{tip2023}, alongside segmentation-based pipelines~\citep{RU-Unet, simlvseg, diag_2023_Batool} for volumetric assessment.
With the advent of deep learning, direct video regression became the mainstream paradigm. The seminal EchoNet-Dynamic~\citep{EchoNet-Dynamic} established the efficacy of 3D CNNs, an approach subsequently extended to diverse clinical scenarios~\citep{EchoNet-Pediatric, tmi_2023_semilvef}. To better capture spatiotemporal dependencies, advanced architectures like Transformers~\citep{miccai_2021_UVT, miccai_coreecho} and Graph Neural Networks~\citep{Miccai_2022_echognn, tim_2025_fhfa} were introduced. Most recently, large-scale foundation models~\citep{EchoClip, panecho} have emerged, offering generalized representations for cardiac analysis.

However, despite these advancements, the underlying paradigm remains predominantly deterministic. Existing methods typically model the conditional expectation $\mathbb{E}[y|\mathbf{x}]$, treating physiological estimation as a fixed point-mapping task. This formulation overlooks the aleatoric uncertainty inherent in noisy ultrasound imagery. Our work departs from this convention by reframing the task as a probabilistic inverse problem, utilizing a generative approach to recover the full posterior distribution.

\subsection{Score-Based Generative Model for Prediction}
Score-based generative models learn the data distribution's score function---the gradient of its log-probability density---to guide generation~\citep{song2019generative, efron2011tweedie, sohl2015deep}. 
This score is effectively learned via score matching~\citep{hyvarinen2005estimation} and its more robust variant, denoising score matching~\citep{vincent2011connection}. 
Early models improved sample quality using multi-scale noise~\citep{song2020sliced}, which, along with Denoising Diffusion Probabilistic Models (DDPMs)~\citep{ho2020denoising, song2020improved}, was unified under a stochastic differential equation (SDE) framework~\citep{song2021scorebased}. 
This framework revealed a deterministic generation process via an ordinary differential equation (ODE), enabling faster sampling with numerical solvers~\citep{DDIM2021, lu2022dpm, lu2025dpm} and maximum likelihood training~\citep{song2021maximum}. 

While originally designed for generation, the potential of diffusion models for discriminative tasks and regression is gaining attention. Unlike implicit models (e.g., GANs) or explicit density models (e.g., VAEs), diffusion models offer stable training and high-quality mode coverage. 
Recent studies have successfully adapted diffusion priors for time series imputation~\citep{tashiro2021csdi}, trajectory forecasting~\citep{zhang2023adding}. Specifically, conditional diffusion models allow for the controlled generation of a target variable $y$ given input covariates $\mathbf{x}$. By learning the conditional score function $\nabla_{y} \log p(y|\mathbf{x})$, these models effectively convert regression into a conditional generation task. 
Inspired by these advances, we propose MCSDR, specifically tailoring the conditional score-based framework to handle the multimodal heterogeneity of clinical data for robust estimation.

\section{Method}
\label{sec:method}

% In this section, we elaborate on the proposed Multimodal Conditional Score-based Diffusion model for Regression (MCSDR). We begin by formulating LVEF estimation as a probabilistic inverse problem (Sec.~\ref{sec:problem_formulation}). We then describe the generative diffusion framework (Sec.~\ref{sec:diffusion_framework}) and the specific architecture of our Multimodal Conditional Score Network (MCSN), designed to effectively fuse heterogeneous clinical data (Sec.~\ref{sec:mcsn}). Finally, we outline the training objective and the inference procedure (Sec.~\ref{sec:inference}).

In this section, we articulate the proposed framework for solving the LVEF estimation inverse problem. We first rigorously define the probabilistic objective (Sec.~\ref{sec:problem_formulation}). We then detail the score-based generative dynamics used to traverse the solution manifold (Sec.~\ref{sec:diffusion_framework}) and the Multimodal Conditional Score Network (MCSN) designed for heterogeneous prior fusion (Sec.~\ref{sec:mcsn}). Finally, we describe the inference procedure for posterior recovery (Sec.~\ref{sec:inference}).

% --- Figure 1: Overview ---
\begin{figure*}[t!]
    \centering
    \includegraphics[width=\linewidth]{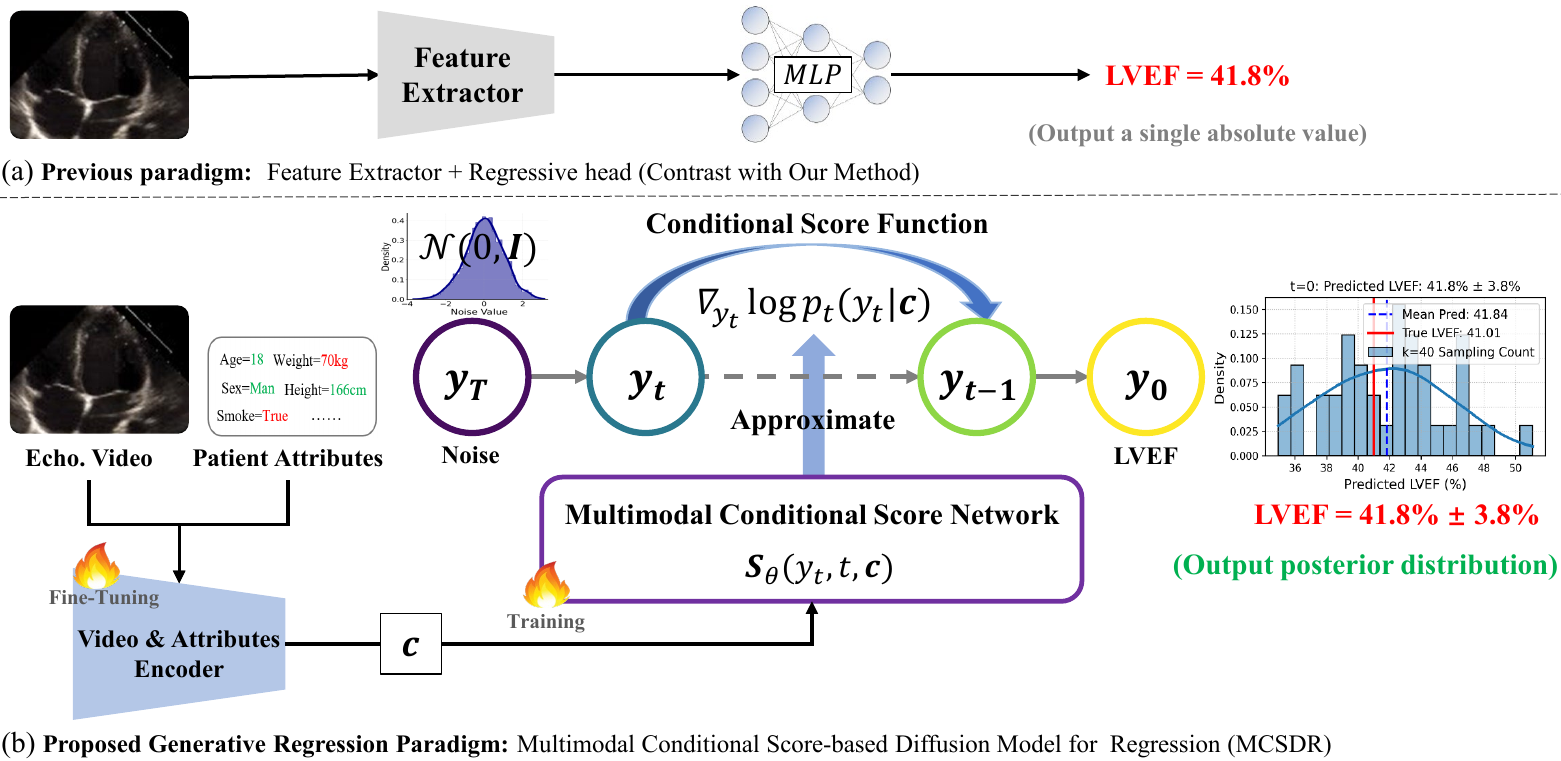} 
    \caption{\textbf{Paradigm shift from deterministic to generative regression.} (a) Conventional methods minimize MSE loss to predict a single point estimate (conditional mean), often failing to capture ambiguity. (b) Our proposed MCSDR framework treats LVEF estimation as a probabilistic inverse problem. It fuses the echocardiogram video with patient attributes to learn a conditional score function. This function guides a reverse diffusion process to generate the full posterior distribution $p(y|\mathbf{x}, \mathbf{c})$.}
    \label{fig:overview}
\end{figure*}
\label{fig:overview}

\subsection{Problem Formulation}
\label{sec:problem_formulation}

Let $y \in \mathcal{Y} \subset \mathbb{R}$ denote the scalar physiological metric (LVEF), and $\mathbf{x} \in \mathcal{X}$ denote the observed echocardiogram video. The imaging process can be conceptually modeled as a forward operator $\mathcal{F}$:
\begin{equation}
    \mathbf{x} = \mathcal{F}(y; \mathbf{z}) + \eta,
\end{equation}
where $\mathbf{z}$ represents unobserved latent factors (e.g., view angle, heart geometry) and $\eta$ denotes measurement noise (e.g., speckle). The goal of LVEF estimation is to invert this process to recover $y$. 

However, this inverse problem is fundamentally \textit{ill-posed}. Due to information loss in $\mathcal{F}$ (projection from 3D motion to 2D planes) and the stochastic nature of $\eta$, the mapping $\mathbf{x} \mapsto y$ is one-to-many. Deterministic regression, which minimizes $\|\hat{y} - y\|^2$, inherently targets the conditional expectation $\mathbb{E}[y|\mathbf{x}]$. In multimodal distributions—common in pathological cases—the mean may fall in a low-density region of the true posterior, yielding a implausible estimate.

To address this, we formulate the task as estimating the full conditional posterior distribution $p(y | \mathbf{x}, \mathbf{a})$. Here, $\mathbf{a} \in \mathcal{A}$ denotes patient demographic attributes (e.g., age, sex). From a Bayesian perspective, these attributes serve as \textbf{informative priors}. While the visual likelihood $p(\mathbf{x}|y)$ constrains the estimate based on motion, the demographic prior $p(y|\mathbf{a})$ constrains the solution space to a physiologically probable range, effectively regularizing the ill-posed inversion:
\begin{equation}
    p(y | \mathbf{x}, \mathbf{a}) \propto p(\mathbf{x} | y, \mathbf{a}) p(y | \mathbf{a}).
\end{equation}
Our framework, MCSDR, implicitly learns this posterior using a score-based generative approach.

% \subsection{Score-Based Generative Regression}
% \label{sec:diffusion_framework}
% To model the continuous distribution of the scalar target $y$, we leverage the score-based generative modeling framework~\citep{song2021scorebased}. This framework involves two coupled stochastic processes: a forward process that diffuses data into noise, and a reverse process that generates data from noise.

% --- Figure 2: Process ---
\begin{figure}[t!]
    \centering
    \includegraphics[width=\linewidth]{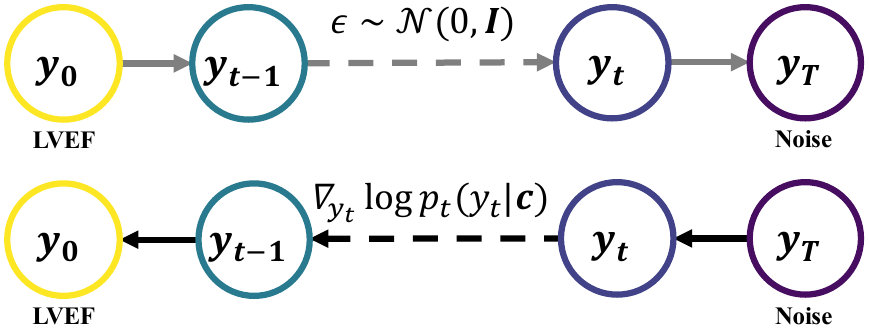} 
    \caption{\textbf{The stochastic generative trajectory.} Top: The forward diffusion process gradually perturbs the scalar LVEF label $y_0$ into Gaussian noise $y_T$. Bottom: The reverse generative process, governed by the learned conditional score function, iteratively solves the inverse problem to reconstruct $y_0$ from noise, conditioned on clinical evidence.}
    \label{fig:diffusion_process}
\end{figure}

\subsection{Score-Based Generative Dynamics}
\label{sec:diffusion_framework}

We leverage the continuous-time diffusion framework~\citep{song2021scorebased} to transform the regression task into a generative process. This involves two coupled stochastic differential equations (SDEs).

\paragraph{Forward Perturbation (Prior Construction)}
The forward process progressively diffuses the ground-truth physiological signal $y_0$ into noise, independent of conditioning variables. It is governed by the Variance Preserving (VP) SDE:
\begin{equation}
    dy_t = -\frac{1}{2} \beta(t)y_t dt + \sqrt{\beta(t)} dw, \quad t \in [0, T],
\end{equation}
where $w$ is a standard Wiener process and $\beta(t)$ is a noise schedule. As $t \rightarrow T$, the distribution of $y_t$ converges to a standard Gaussian prior $p_T(y) \approx \mathcal{N}(0, 1)$, effectively wiping out the original signal information.

\paragraph{Reverse Restoration (Manifold Recovery)}
The generative process corresponds to reversing the diffusion time from $T$ to $0$. It is governed by the reverse-time SDE, or equivalently for deterministic sampling, the Probability Flow ODE~\citep{song2021scorebased}:
\begin{equation}
    dy_t = \left[ -\frac{1}{2} \beta(t)y_t - \frac{1}{2} \beta(t)\nabla_{y_t} \log p_t(y_t|\mathbf{c}) \right] dt.
    \label{eq:ode_reverse}
\end{equation}
where $\mathbf{c}$ denotes the \textbf{unified multimodal embedding} derived from the raw inputs $\mathbf{x}$ and $\mathbf{a}$ (i.e., $\mathbf{c} = \text{Encoder}(\mathbf{x}, \mathbf{a})$).
It is worth noting that unlike the reverse-time SDE which utilizes a score coefficient of $-\beta(t)$, the deterministic ODE formulation scales the score term by $-\frac{1}{2}\beta(t)$ to match the marginal densities without stochastic injection. The core term here is the \textbf{conditional score function} $\nabla_{y_t} \log p_t(y_t|\mathbf{c})$. Geometrically, this score represents a vector field (in 1D, a gradient) that points towards high-density regions of the conditional data distribution. By following this gradient field, we can traverse from the noise domain back to the physiological manifold. Since the true score is unknown, we approximate it using a time-dependent neural network $S_\theta(y_t, t, \mathbf{c})$, described below.

% \subsection{Multimodal Conditional Score Network (MCSN)}
% \label{sec:mcsn}

% --- Figure 3: MCSN ---
\begin{figure*}[t!]
    \centering
    \includegraphics[width=\linewidth]{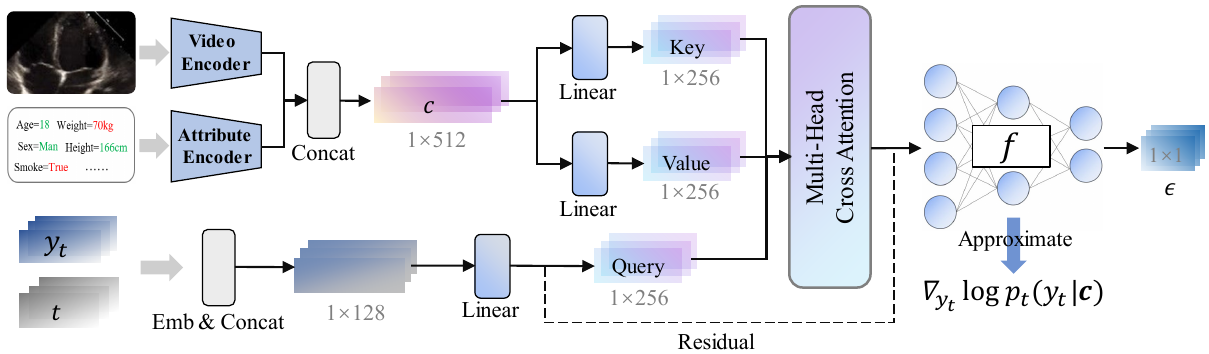}
    \caption{\textbf{Architecture of the Multimodal Conditional Score Network (MCSN).} The network estimates the gradient field necessary for the generative process. To effectively fuse heterogeneous data, we employ dual encoders: a \textbf{Video Encoder} for spatiotemporal echocardiogram features and an \textbf{Attribute Encoder} for tabular clinical priors (e.g., age, weight). These features are concatenated and projected to form the Keys ($\mathbf{K}$) and Values ($\mathbf{V}$) in the cross-attention mechanism. This design allows the model to dynamically condition the denoising of the LVEF state (Query $\mathbf{Q}$) on specific patient contexts.}
    \label{fig:mcsn_arch}
\end{figure*}

\subsection{Multimodal Conditional Score Network (MCSN)}
\label{sec:mcsn}

The accuracy of the inverse solution hinges on the quality of the estimated gradient field. We propose the MCSN architecture to effectively fuse the heterogeneous modalities—high-dimensional spatiotemporal video $\mathbf{x}$ and low-dimensional tabular priors $\mathbf{a}$—to condition the score estimation.

\paragraph{Heterogeneous Feature Encoding}
We employ a dual-branch encoder design to handle the distinct nature of the inputs:
\begin{itemize}
    \item \textbf{Visual Encoder:} The echocardiogram video $\mathbf{x}$ is processed by a pre-trained spatiotemporal backbone~\citep{miccai_coreecho} to extract a dense visual embedding $\mathbf{h}_{vis} \in \mathbb{R}^{d_v}$. This vector encapsulates the motion patterns and structural deformations essential for the likelihood term $p(\mathbf{x}|y)$.
    \item \textbf{Attribute Encoder:} Tabular demographic attributes $\mathbf{a}$ are processed via a Multi-Layer Perceptron (MLP) to project discrete and continuous variables into a latent semantic space $\mathbf{h}_{attr} \in \mathbb{R}^{d_a}$. This embedding encodes the Bayesian prior $p(y|\mathbf{a})$, providing demographic constraints.
\end{itemize}

\paragraph{Dynamic Gradient Modulation via Attention}
To integrate these features, we treat the score estimation as a conditional modulation problem. The noisy state $y_t$ is first embedded with sinusoidal time features to form the Query vector $\mathbf{Q}$. The visual and attribute embeddings are concatenated to form the context $\mathbf{c} = [\mathbf{h}_{vis}; \mathbf{h}_{attr}]$, which is projected into Keys $\mathbf{K}$ and Values $\mathbf{V}$.
We employ a Multi-Head Cross-Attention mechanism:
\begin{equation}
    \text{Attention}(\mathbf{Q}, \mathbf{K}, \mathbf{V}) = \text{softmax} \left(\frac{\mathbf{Q}\mathbf{K}^T}{\sqrt{d_k}}\right) \mathbf{V}.
\end{equation}
This mechanism allows the network to dynamically weigh the evidence. For instance, in frames where visual features are ambiguous (e.g., severe noise), the attention mechanism can attend more heavily to the demographic priors ($\mathbf{h}_{attr}$) to guide the gradient computation, thereby stabilizing the inverse solution.

% \subsection{Training Objective}
% \label{sec:training}
% Following the principle of Denoising Score Matching, estimating the score function is equivalent to training a noise predictor $\boldsymbol{\epsilon}_\theta$. We adopt the simplified objective used in DDPM~\citep{ho2020denoising}, which minimizes the Mean Squared Error (MSE) between the actual noise $\boldsymbol{\epsilon}$ added in the forward process and the network's prediction:
% \begin{equation}
%     \mathcal{L}(\theta) = \mathbb{E}_{t, y_0, \boldsymbol{\epsilon}} \left[ \left\| \boldsymbol{\epsilon} - \boldsymbol{\epsilon}_\theta(y_t, t, \mathbf{c}) \right\|^2 \right] . \tag{6}
% \end{equation}
% The network is trained end-to-end. To preserve the generalization capability of the pre-trained visual backbone, we apply a differential learning rate strategy during optimization.

\subsection{Training Objective}
\label{sec:training}
Following the standard denoising score matching formulation~\citep{ho2020denoising}, we train the network to predict the noise $\boldsymbol{\epsilon}$ added to $y_0$. The objective minimizes the MSE between the predicted noise and the actual noise:
\begin{equation}
    \mathcal{L}(\theta) = \mathbb{E}_{t, y_0, \mathbf{c}, \boldsymbol{\epsilon}} \left[ \left\| \boldsymbol{\epsilon} - \boldsymbol{\epsilon}_\theta(y_t, t, \mathbf{c}) \right\|^2 \right].
\end{equation}
Crucially, although we use an MSE loss on the \textit{noise}, this is mathematically equivalent to minimizing the Fisher Divergence between the data distribution and the model distribution. This differs fundamentally from direct MSE regression on $y$, as it learns the structure of the distribution (the score) rather than just its mean.

\subsection{Inference and Solution Analysis}
\label{sec:inference}

During inference, our goal is to approximate the posterior $p_\theta(y | \mathbf{c})$. Since the exact analytical form is intractable, we employ a Monte Carlo sampling strategy.

\paragraph{Deterministic Sampling via DDIM}
To ensure computational efficiency, we utilize the Denoising Diffusion Implicit Model (DDIM) solver~\citep{DDIM2021}. By setting the variance parameter $\eta=0$, the reverse process becomes deterministic. This allows us to solve the Probability Flow ODE (Eq.~\ref{eq:ode_reverse}) with a reduced number of steps (e.g., $\tau=10$), mapping a specific noise sample $y_T$ uniquely to a predicted $y_0$.

\paragraph{Monte Carlo Posterior Approximation}
To capture the full solution space of the ill-posed inverse problem, we sample $K$ independent trajectories starting from standard Gaussian noise $y_T^{(k)} \sim \mathcal{N}(0, 1)$ for $k=1, \dots, K$. Each trajectory represents a plausible hypothesis consistent with the conditional evidence.
The final estimate is derived from the empirical statistics of these hypotheses:
\begin{equation}
    \hat{y} = \frac{1}{K} \sum_{k=1}^K \hat{y}_0^{(k)}, \quad \sigma = \sqrt{\frac{1}{K-1} \sum_{k=1}^K (\hat{y}_0^{(k)} - \hat{y})^2}.
\end{equation}
Here, $\hat{y}$ represents the ensemble prediction (approximating the posterior mean), while $\sigma$ characterizes the dispersion of the solution space. A large $\sigma$ indicates a flat or multimodal posterior landscape, signaling high ambiguity in the inverse problem (e.g., conflicting visual evidence), whereas a small $\sigma$ indicates a sharp, well-posed solution.

\section{Experiments}
\label{sec:experiments}

\subsection{Datasets and Experimental Setup}
\label{subsec:datasets}

We evaluate our framework on three public echocardiography datasets, chosen to represent varying degrees of clinical complexity and data modality availability.

\textbf{EchoNet-Dynamic~\cite{EchoNet-Dynamic}.} A large-scale dataset comprising 10,030 apical-4-chamber (A4C) echocardiogram videos. While it includes textual reports, these contain direct measurements (e.g., EDV, ESV) that would cause label leakage if used as input. Therefore, we utilize this dataset exclusively for \textbf{vision-only} experiments to benchmark the visual encoding capability of our model.

\textbf{EchoNet-Pediatric~\cite{EchoNet-Pediatric}.} A pediatric dataset containing 3,682 A4C videos from patients aged 0--18 years. This dataset presents significant challenges due to the high physiological variability across developmental stages. It includes tabular demographic attribute details (e.g., age, sex, weight, height), which we utilize as patient demographics for multimodal conditioning.

\textbf{CAMUS~\cite{CAMUS}.} A dataset containing 500 patients with both A2C and A4C views. It is characterized by variable image quality and potential artifacts, representing a challenging inverse problem scenario. Similar to EchoNet-Pediatric, we use the provided patient demographics as tabular priors.

\textbf{Evaluation Metrics.} 
% We employ three standard deterministic metrics: Mean Absolute Error (MAE), Root Mean Squared Error (RMSE), and the Coefficient of Determination ($R^2$). Additionally, to rigorously evaluate the quality of the generated posterior distributions, we report the Continuous Ranked Probability Score (CRPS). CRPS generalizes MAE to probabilistic forecasts, measuring how well the predicted distribution aligns with the ground truth observation. Lower values indicate better calibration and accuracy for MAE, RMSE, and CRPS.

We employ standard deterministic metrics: Mean Absolute Error (MAE), Root Mean Squared Error (RMSE), and the Coefficient of Determination ($R^2$). 
Crucially, to evaluate the model's ability to recover the true data distribution rather than just the conditional mean, we report the Continuous Ranked Probability Score (CRPS). CRPS generalizes MAE to probabilistic forecasts, measuring the distance between the predicted posterior cumulative distribution function (CDF) and the empirical step function of the ground truth. Lower CRPS indicates better distributional fidelity.

\subsection{Implementation Details}
% Implemented in PyTorch on NVIDIA RTX 4090 GPUs, our framework employs a unified data processing pipeline. Input video frames ($112 \times 112$) undergo minimal geometric augmentation via \texttt{albumentations}, specifically zero-padding to $124 \times 124$ followed by random cropping during training and deterministic center cropping for evaluation. Patient demographics, where applicable (EchoNet-Pediatric and CAMUS), are normalized or one-hot encoded and concatenated with visual embeddings to form the condition $\mathbf{c}$; The architecture utilizes a pre-trained Uniformer-Small~\cite{miccai_coreecho} backbone, which is fine-tuned using a differential learning rate strategy (10\% of the base LR), while the randomly initialized score estimation head is optimized via AdamW~\cite{adamw} (base LR $3 \times 10^{-4}$, weight decay $1 \times 10^{-5}$). We strictly adhere to the official splits for EchoNet-Dynamic and CAMUS. For EchoNet-Pediatric, we constructed a fixed 8:1:1 split to ensure reproducible comparisons, as the original study employed random cross-validation\cite{EchoNet-Pediatric}. For remaining technical specifics, please refer to our publicly released code.

MCSDR is implemented in PyTorch. The video encoder utilizes a pre-trained Uniformer-Small~\cite{miccai_coreecho} backbone, which is fine-tuned using a differential learning rate strategy (10\% of the base LR). The diffusion process is governed by a Variance Preserving (VP) SDE with a linear $\beta(t)$ schedule ranging from $10^{-4}$ to $0.02$ over $T=1000$ discrete training steps.
During training, we optimize the denoising objective (Eq. 6) using AdamW (LR $3 \times 10^{-4}$, weight decay $1 \times 10^{-5}$) on NVIDIA RTX 4090 GPUs. 
During inference, we approximate the posterior using the deterministic DDIM solver ($\eta=0$) with $\tau=10$ steps to balance precision and latency. The final prediction is the ensemble mean of $K=40$ stochastic trajectories sampled from the standard Gaussian prior.
For remaining technical specifics, please refer to our publicly released code (\url{https://github.com/lvmarch/MCSDR.git}).

\subsection{Comparison with State-of-the-Art Methods}
\label{sec:comparison_sota}

To rigorously evaluate the proposed generative paradigm, we compare MCSDR against a comprehensive set of baselines. These include traditional deterministic regression models employing diverse backbones—CNNs (EchoNet~\cite{EchoNet-Dynamic}, FHFA-MTL~\cite{tim_2025_fhfa}), Transformers (UVT~\cite{miccai_2021_UVT}, EchoCoTr~\cite{miccai_2022_echocotr}, CoReEcho~\cite{miccai_coreecho}), and Graph Neural Networks (EchoGNN~\cite{Miccai_2022_echognn}, EchoGraphs~\cite{Miccai_2022_echograph})—as well as recent large-scale medical foundation models (EchoClip~\cite{EchoClip}, PanEcho~\cite{panecho}).

\textbf{Benchmark on Large-Scale Adult Echocardiography.}
Table~\ref{tab:echonet_dynamic} presents the results on the EchoNet-Dynamic test set. To ensure a strictly fair comparison and prevent potential label leakage from the dataset's derived text attributes, our MCSDR model is evaluated in a vision-only mode.
Despite utilizing only visual inputs, MCSDR establishes a new state-of-the-art, achieving an MAE of 3.76. Notably, it outperforms the recent foundation model PanEcho (MAE 5.35) and the latest multitask regression framework FHFA-MTL (MAE 3.89). 
This result suggests that the performance bottleneck in current LVEF estimation may not lie in the feature extractor's capacity, but rather in the regression paradigm itself. By modeling the target as a distribution rather than a point estimate.

% --- Table 1: Comparison on EchoNet-Dynamic (Vision-Only) ---
\begin{table}[t!]
\centering
\small
\setlength{\tabcolsep}{4pt} 
\caption{Comparison with SOTA methods on the \textbf{EchoNet-Dynamic}. ``--'' indicates unavailable results. ``*'' indicates reproducible source code.}
% \begin{tabular}{l|l|ccc}
\begin{tabular*}{\columnwidth}{l|l@{\extracolsep{\fill}}ccc}
\toprule
\textbf{Type} & \textbf{Method} & \begin{tabular}[c]{@{}c@{}}\textbf{MAE} \small{$\downarrow$}\end{tabular} & \begin{tabular}[c]{@{}c@{}}\textbf{RMSE} \small{$\downarrow$}\end{tabular} & \begin{tabular}[c]{@{}c@{}}\textbf{R\textsuperscript{2}} \small{$\uparrow$}\end{tabular} \\
\midrule
\multirow{2}{*}{\begin{tabular}[c]{@{}c@{}}Foundation\\Model\end{tabular}} & EchoClip\cite{EchoClip} & 7.14 & - & - \\
& PanEcho\cite{panecho} & 5.35 & - & - \\
\midrule
\multirow{7}{*}{\begin{tabular}[c]{@{}c@{}}Deterministic \\Regression\end{tabular}} 
& *EchoNet(32F)\cite{Miccai_2022_echognn} & 4.22 & 5.56 & 0.79 \\ 
& *UVT\cite{miccai_2021_UVT} & 5.95 & 8.38 & 0.52 \\ 
& *EchoGNN\cite{Miccai_2022_echognn} & 4.45 & 5.97 & 0.76 \\ 
& *EchoGraphs\cite{Miccai_2022_echograph} & 4.01 & 5.36 & 0.81 \\
& *EchoCoTr\cite{miccai_2022_echocotr} & 3.95 & 5.17 & 0.82 \\
& Batool et al.\cite{diag_2023_Batool} & 5.74 & 7.73 & 0.60 \\ 
& *CoReEcho\cite{miccai_coreecho} & 3.90 & 5.13 & 0.82 \\
& FHFA-MTL\cite{tim_2025_fhfa} & 3.89 & 5.13 & 0.82 \\
\midrule
\begin{tabular}[c]{@{}c@{}}Generative\\Regression\end{tabular} 
& \begin{tabular}[c]{@{}c@{}}MCSDR\\(Vision only)\end{tabular} 
& \textbf{3.76} & \textbf{4.81} & \textbf{0.84} \\
\bottomrule
\end{tabular*}
\label{tab:echonet_dynamic}
\end{table}

% --- Table 2: Comparison on EchoNet-Pediatric (Multimodal) ---
\begin{table}[t!]
\centering
\small
\setlength{\tabcolsep}{6pt} 
\caption{Comparison with SOTA methods on the \textbf{EchoNet-Pediatric}.}
% \begin{tabular}{l|ccc}
\begin{tabular*}{\columnwidth}{l@{\extracolsep{\fill}}ccc}
\toprule
\textbf{Method} & \textbf{MAE} $\downarrow$ & \textbf{RMSE} $\downarrow$ & \textbf{R\textsuperscript{2}} $\uparrow$ \\
\midrule
% *EchoNet-Ped (A4C)\cite{EchoNet-Pediatric} & 4.15 & 5.70 & 0.70 \\
*UVT\cite{miccai_2021_UVT} & 7.91 & 9.47 & 0.33 \\ 
*EchoGNN\cite{Miccai_2022_echognn} & 5.69 & 7.28 & 0.60 \\ 
*EchoGraphs\cite{Miccai_2022_echograph} & 5.29 & 6.87 & 0.65 \\ 
*EchoCoTr\cite{miccai_2022_echocotr} & 4.82 & 6.26 & 0.71 \\ 
*CoReEcho\cite{miccai_coreecho} & 4.70 & 6.40 & 0.69 \\ 
\midrule
MCSDR (Vision only) & 4.26 & 5.88 & 0.74 \\ 
MCSDR (Vision + Demographis) & \textbf{4.22} & \textbf{5.59} & \textbf{0.77} \\
\bottomrule
\end{tabular*}
\label{tab:echonet_pediatric}
\end{table}

% --- Table 3: Comparison on CAMUS (Multimodal) ---
\begin{table}[t!]
\centering
\small
\setlength{\tabcolsep}{6pt} 
\caption{Comparison with SOTA methods on the \textbf{CAMUS}.}
% \begin{tabular}{l|ccc}
\begin{tabular*}{\columnwidth}{l@{\extracolsep{\fill}}ccc}
\toprule
\textbf{Method} & \textbf{MAE} $\downarrow$ & \textbf{RMSE} $\downarrow$ & \textbf{R\textsuperscript{2}} $\uparrow$ \\
\midrule
UVT\cite{miccai_2021_UVT} & 9.42 & 11.02 & 0.27 \\
EchoGNN\cite{Miccai_2022_echognn} & 8.58 & 9.73 & 0.43 \\
EchoGraphs\cite{Miccai_2022_echograph} & 8.70 & 9.66 & 0.44 \\
EchoCoTr\cite{miccai_2022_echocotr} & 7.38 & 8.99 & 0.51 \\
CoReEcho\cite{miccai_coreecho} & 7.35 & 8.76 & 0.54 \\
\midrule
MCSDR (Vision only) & 6.22 & 8.31 & 0.58 \\
MCSDR (Vision + Demographis) & \textbf{5.73} & \textbf{7.50} & \textbf{0.66} \\
\bottomrule
\end{tabular*}
\label{tab:camus}
\end{table}

\textbf{Benchmark on High-Variability Pediatric Data.}
The EchoNet-Pediatric dataset presents challenges stemming from high physiological variability across developmental stages (ages 0–18). As shown in Table 2, our approach outperforms all baselines, including data-release methods that benefit from extensive pre-training. Notably, incorporating patient demographics as conditional priors further improves $R^2$ from 0.74 to 0.77. This demonstrates that patient-specific demographics are highly beneficial for physiological metric prediction.

\textbf{Benchmark on Small-Scale, Noisy Data.}
The CAMUS dataset represents a classic ill-posed inverse problem scenario, characterized by a small sample size ($N=500$) and variable, often suboptimal, image quality. The results in Table~III provide the most compelling evidence for our multimodal generative formulation.
Deterministic baselines degrade significantly in this setting, with CoReEcho yielding an MAE of 7.35. Our vision-only MCSDR significantly reduces this error to 6.22, demonstrating the data efficiency of the diffusion training objective. 
Most importantly, the fusion of patient demographics brings a substantial performance leap, further reducing the MAE to 5.73. This empirical evidence validates our core hypothesis: when visual evidence is ambiguous due to noise or artifacts, the generative process effectively leverages clinical priors to constrain the solution space, guiding the trajectory toward the ground truth.

\begin{figure}[t!]
\centering
\includegraphics[width=\columnwidth]{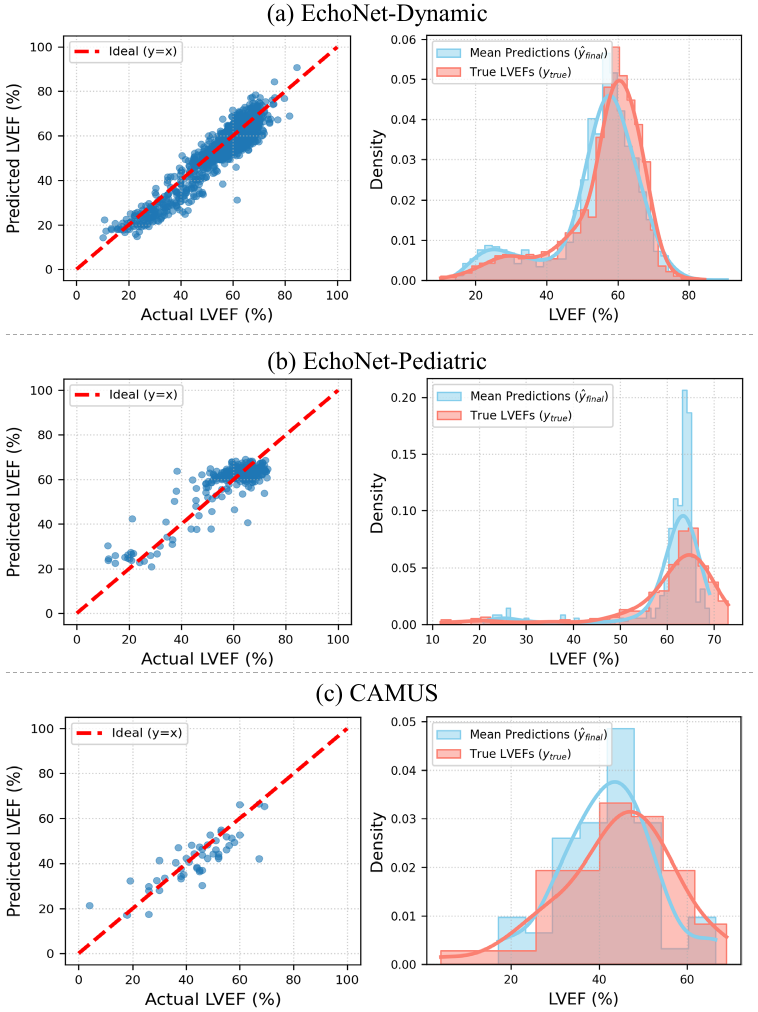}
\caption{Visual analysis of model predictions on (a) CAMUS, (b) EchoNet-Pediatric, and (c) EchoNet-Dynamic test sets. Left: Scatter plots of Predicted vs. True LVEF. Right: Density plots comparing the distribution of predictions (Blue) against the ground truth (Red).}
\label{fig:results_scatter_density}
\end{figure}

\subsection{Visual and Distribution Analysis}
\label{sec:visual_analysis}

To substantiate the quantitative metrics, we provide a qualitative assessment of the model's predictive behavior in Figure~\ref{fig:results_scatter_density}. The visualization consists of scatter plots (Left) illustrating instance-level accuracy and density plots (Right) evaluating population-level distributional fidelity.

\textbf{Instance-Level Alignment.}
On the large-scale EchoNet-Dynamic dataset (Fig.~\ref{fig:results_scatter_density}a), the predictions exhibit remarkable tightness around the ideal identity line ($y=x$), correlating with the high $R^2$ of 0.84. 
For the EchoNet-Pediatric dataset (Fig.~\ref{fig:results_scatter_density}b), despite the high physiological variability inherent in the 0--18 age range, the scatter plot maintains a linear trajectory with minimal outliers. 
Even on the challenging CAMUS dataset (Fig.~\ref{fig:results_scatter_density}c), where image quality is suboptimal, the predictions remain unbiased. The increased dispersion observed in the CAMUS scatter plot correctly reflects the higher aleatoric uncertainty associated with noisy inputs, rather than systematic model failure.

\textbf{Distributional Fidelity.}
A common pitfall in deterministic regression is the tendency to collapse towards the conditional mean, often resulting in a predicted distribution that is narrower than the ground truth (i.e., underestimating the variance of the population). 
In contrast, the density plots in the right column demonstrate that MCSDR preserves the true statistical properties of the target variable. 
As shown in Fig.~\ref{fig:results_scatter_density}a and b, the probability density functions of our mean predictions (Blue) align almost perfectly with the ground truth densities (Red). 
Notably, on EchoNet-Pediatric, the model accurately reproduces the high kurtosis (sharp peak) centered around 65\% LVEF. 
On CAMUS, it successfully captures the broader, more platykurtic distribution. 
This generative consistency confirms that our probabilistic formulation effectively learns the underlying data manifold, ensuring that the generated LVEF values are not only accurate for individual patients but also statistically valid across the cohort.

% --- Table 4: Combined Ablation ---
\begin{table}[t!]
\centering
\small
\caption{Ablation on paradigm effectiveness and computational cost.}
\setlength{\tabcolsep}{3pt}
\begin{tabular*}{\columnwidth}{l@{\extracolsep{\fill}}ccccc}
\toprule
\textbf{Method} & \textbf{MAE} & \textbf{RMSE} & \textbf{R\textsuperscript{2}} & \textbf{FLOPs} & \textbf{Time} \\
\midrule
Deterministic Reg. & 5.26 & 6.85 & 0.65 & \textasciitilde13.8G & \textbf{0.01s} \\
MCSDR (DDPM) & 4.20 & 5.69 & 0.75 & \textasciitilde69.1G & 1.17s \\
\textbf{MCSDR (DDIM)} & \textbf{4.13} & \textbf{5.55} & \textbf{0.77} & \textasciitilde14.4G & 0.02s \\
\bottomrule
\end{tabular*}
\label{tab:ablation_paradigm_perf}
\end{table}

\subsection{Ablation Studies}

\subsubsection{Generative vs. Deterministic Regression}
We first validate the core hypothesis that formulating regression as a generative task is beneficial. We compare MCSDR against a baseline that uses the exact same encoder backbone (Uniformer + Attribute MLP) but is trained with a standard regression head and MSE loss.
As shown in Table~\ref{tab:ablation_paradigm_perf}, the generative formulation (using DDIM) improves MAE from 5.26 to 4.13 on EchoNet-Pediatric. This supports our premise that learning the gradient field of the data distribution allows for more robust inference than directly regressing the conditional mean, particularly for complex inverse problems.

\subsubsection{Sampler Selection: DDPM vs. DDIM}
\begin{figure}[t]
\centering
\includegraphics[width=\columnwidth]{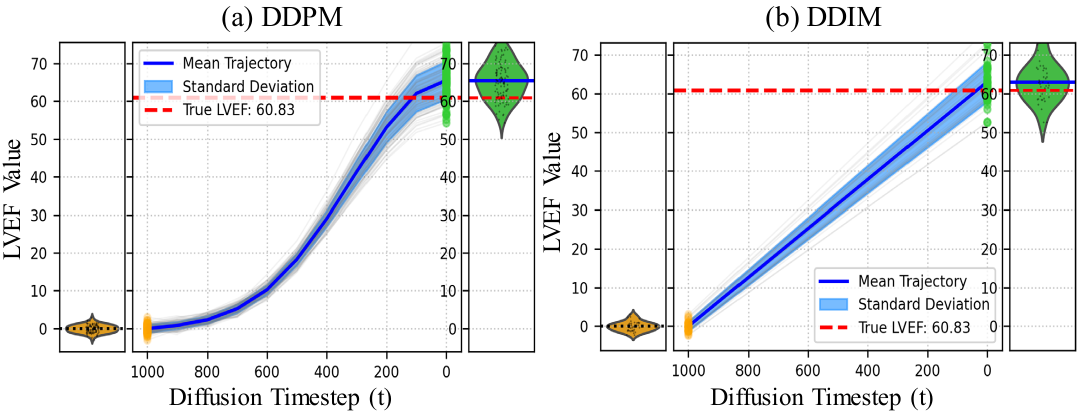}
\caption{Visual comparison of sampling trajectories. Top: Stochastic DDPM path ($T=1000$). Bottom: Deterministic DDIM path ($\tau=10$). The nearly linear trajectory of the ODE-based DDIM allows for efficient sampling with fewer steps.}
\label{fig:sampler_paths}
\end{figure}

We compare two sampling strategies for solving the reverse SDE/ODE. Table~\ref{tab:ablation_paradigm_perf} highlights the trade-off. While the stochastic DDPM sampler ($T=1000$) achieves competitive accuracy, it incurs high latency (1.17s per sample). The deterministic DDIM sampler ($\eta=0, \tau=10$) achieves comparable or slightly better accuracy (MAE 4.13) but with drastically reduced inference time (0.02s). This efficiency is attributed to the semi-linear nature of the Probability Flow ODE trajectory (visualized in Figure~\ref{fig:sampler_paths}), which allows for larger integration steps without significant discretization error.

\subsubsection{Architecture Effectiveness}
We further validate the design of the MCSN by comparing it with a simple MLP predictor and a Transformer Decoder. Table~\ref{tab:ablation_mcsn} shows that our dual-encoder MCSN architecture achieves the lowest CRPS (3.01) and MAE with significantly fewer parameters (0.69M) compared to the Transformer Decoder (2.27M). This indicates that the specific design for fusing spatiotemporal features with tabular attributes is parameter-efficient.

% \begin{table}[h!]
% \centering
% \small
% \caption{Ablation on network architecture.}
% \setlength{\tabcolsep}{4pt}
% \begin{tabular*}{\columnwidth}{l@{\extracolsep{\fill}}cccc}
% \toprule
% \textbf{Architecture} & \textbf{Params} & \textbf{MAE} $\downarrow$ & \textbf{R\textsuperscript{2}} $\uparrow$ & \textbf{CRPS} $\downarrow$ \\
% \midrule
% MLP & 0.23M & 4.27 & 0.74 & 3.10 \\
% Transformer Decoder & 2.27M & 4.30 & 0.74 & 3.04 \\
% \textbf{MCSN (Ours)} & \textbf{0.69M} & \textbf{4.13} & \textbf{0.77} & \textbf{3.01} \\
% \bottomrule
% \end{tabular*}
% \label{tab:ablation_mcsn}
% \end{table}

% --- Table 4: Ablation on the MCSN Architecture (Single Column) ---
\begin{table}[t!]
\centering
\small
\setlength{\tabcolsep}{4pt} 
\caption{Ablation study on the noise predictor architecture.}
% \begin{tabular}{l|cccc|c}
\begin{tabular*}{\columnwidth}{l@{\extracolsep{\fill}}|cccc|c}
\toprule
\textbf{Architecture} & \textbf{MAE} $\downarrow$ & \textbf{RMSE} $\downarrow$ & \textbf{R\textsuperscript{2}} $\uparrow$ & \textbf{CRPS} $\downarrow$ & \textbf{Params.} \\
\midrule
\multicolumn{6}{c}{\textit{Results using DDPM Sampler (T=1000 steps)}} \\
\cmidrule(lr){1-6}
MLP & 4.73 & 6.48 & 0.68 & 3.71 & 0.23M \\
T-Decoder & 4.43 & \textbf{5.55} & \textbf{0.77} & 3.28 & 2.27M \\
MCSN & \textbf{4.20} & 5.69 & 0.75 & \textbf{3.04} & \textbf{0.69M} \\
\midrule
\multicolumn{6}{c}{\textit{Results using DDIM Sampler (S=10 steps)}} \\
\cmidrule(lr){1-6}
MLP & 4.27 & 5.79 & 0.74 & 3.10 & 0.23M \\
T-Decoder & 4.30 & 5.87 & 0.74 & 3.04 & 2.27M \\
MCSN & \textbf{4.13} & \textbf{5.55} & \textbf{0.77} & \textbf{3.01} & \textbf{0.69M} \\
\bottomrule
\end{tabular*}
\label{tab:ablation_mcsn}
\end{table}

\begin{figure}[]
\centering
\includegraphics[width=\columnwidth]{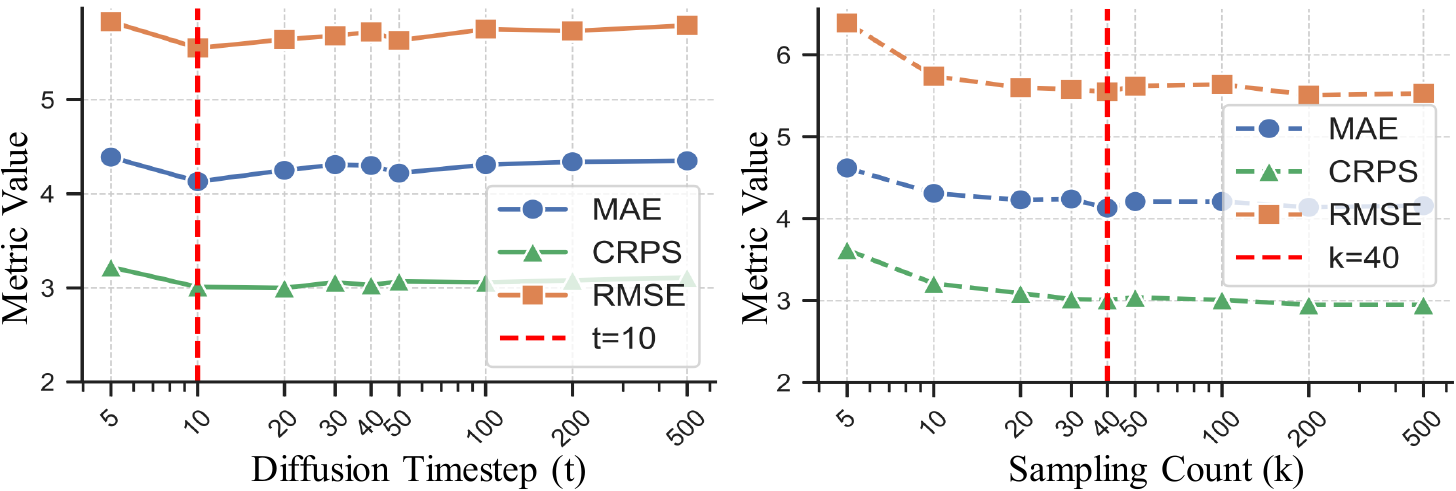}
\caption{The impact of key MCSDR hyperparameters.}
\label{fig:hyperparameter_ablation}
\end{figure}

\subsubsection{Impact of Key Hyperparameters}
We analyze the impact of the DDIM sampling steps ($t$) and the posterior sampling count ($K$) on the EchoNet-Pediatric dataset.
Figure~\ref{fig:hyperparameter_ablation} (left) shows that model performance (MAE, RMSE, CRPS) is optimal at only 10 sampling steps. Using more steps provides no benefit and increases computational cost.
Figure~\ref{fig:hyperparameter_ablation} (right) shows that performance rapidly improves as the sampling count $K$ increases. The metrics stabilize around $K=40$, indicating that a stable posterior approximation has been achieved.
Based on this, we select $t=10$ and $K=40$ as our optimal configuration, achieving the best balance of accuracy and efficiency.

% --- Figure 7: Posterior Adaptability ---
\begin{figure*}[t]
    \centering
    \includegraphics[width=\textwidth]{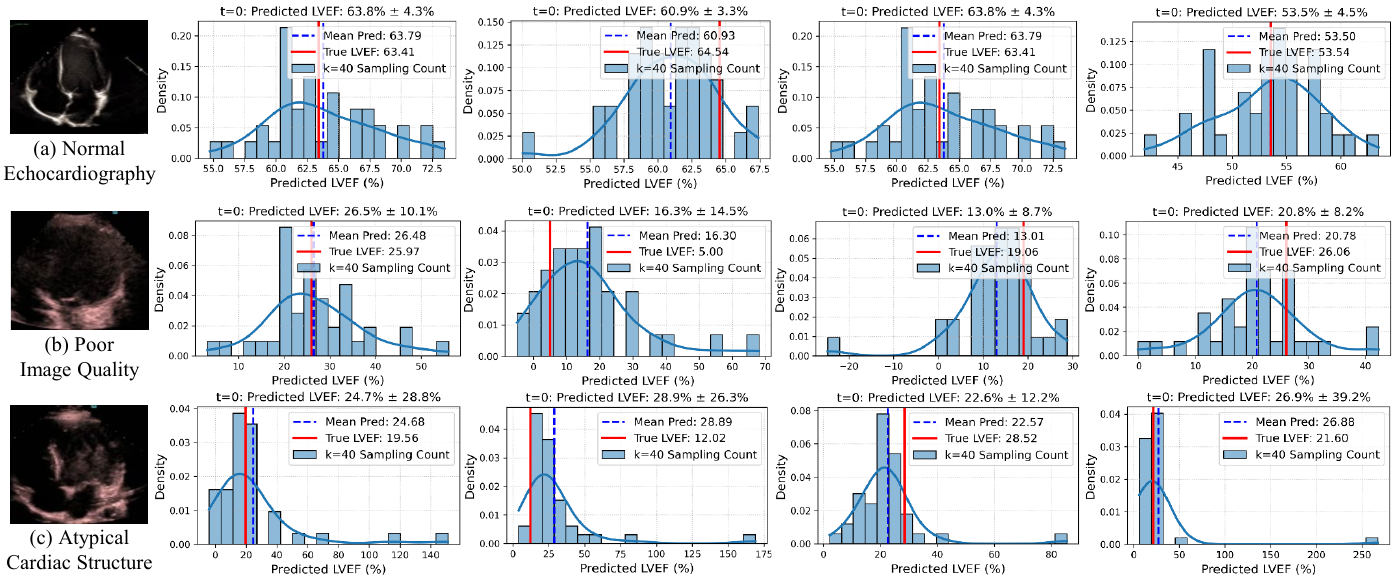}
    \caption{\textbf{Posterior Adaptability.} The model's output distribution adapts to clinical ambiguity. (a) High-quality inputs yield sharp posteriors. (b-c) Noisy or atypical inputs result in dispersed posteriors.}
    \label{fig:posterior_adaptability}
\end{figure*}

% --- Figure 8: Trajectory Dynamics ---
\begin{figure*}[t]
    \centering
    \includegraphics[width=\textwidth]{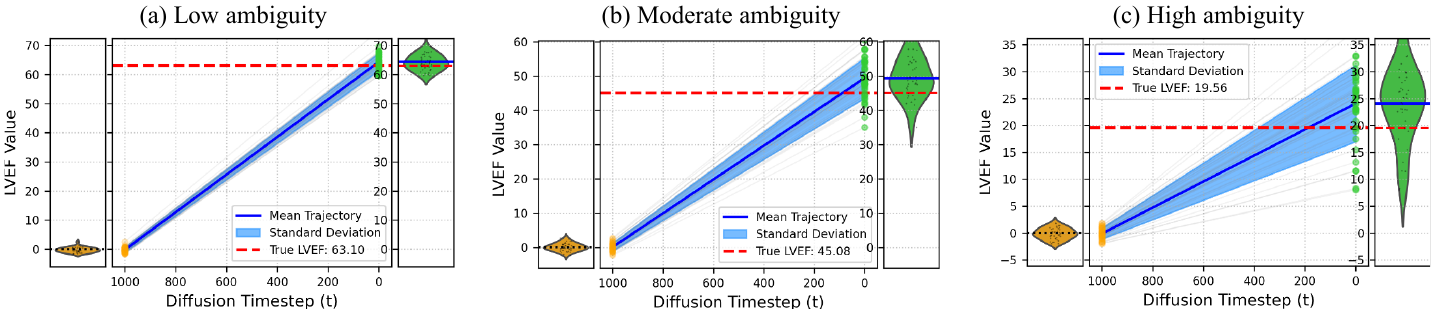}
    \caption{\textbf{Trajectory Coherence Analysis.} The DDIM trajectories ($t=T \to 0$) reveal the geometry of the optimization landscape. (a) Coherent convergence implies a well-posed solution driven by strong visual evidence. (c) Trajectory dispersion faithfully reflects the ambiguity (ill-posedness) of the input, serving as an intrinsic reliability indicator.}
    \label{fig:trajectory_dynamics}
\end{figure*}

\subsection{Qualitative Analysis of Generative Dynamics}
\label{sec:generative_dynamics}
Beyond deterministic accuracy, a key property of our score-based framework is its transparency. By visualizing the generative trajectories solving the probability flow ODE, we can interpret how the model navigates the inverse solution space given different levels of clinical ambiguity.

\textbf{Posterior Adaptability to Input Ambiguity.}
Figure~\ref{fig:posterior_adaptability} demonstrates the model's ability to adapt the shape of its predicted posterior distribution $p_\theta(y|\mathbf{c})$ according to the quality of the visual evidence.
For standard, high-quality echocardiograms (Fig.~\ref{fig:posterior_adaptability}a), the model generates a sharp, unimodal distribution (std dev $\approx 4\%$), indicating a high-confidence consensus among the sampled hypotheses.
In contrast, when the input is degraded by noise (Fig.~\ref{fig:posterior_adaptability}b) or presents atypical cardiac anatomy (Fig.~\ref{fig:posterior_adaptability}c), the posterior naturally dilates (std dev $> 10\%$). This phenomenon confirms that the variance in our regression head is not random noise, but a calibrated response to the \textit{ill-posedness} of the specific input sample.

\textbf{Trajectory Coherence and Bifurcation.}
To further understand the origin of this variance, we visualize the DDIM sampling trajectories in Figure~\ref{fig:trajectory_dynamics}.
In low-uncertainty cases (Fig.~\ref{fig:trajectory_dynamics}a), we observe strong \textbf{trajectory coherence}: despite starting from diverse random noise states $y_T$, the conditional score function acts as a powerful attractor, guiding all paths to converge rapidly into a tight bundle at $t=0$. This indicates a steep, well-defined optimization landscape.
Conversely, in high-uncertainty scenarios (Fig.~\ref{fig:trajectory_dynamics}c), the trajectories exhibit \textbf{bifurcation} or fanning. The score fields in these regions are flatter or multi-modal, allowing different noise initializations to drift towards disparate solutions.

\begin{figure}[t]
    \centering
    \includegraphics[width=\columnwidth]{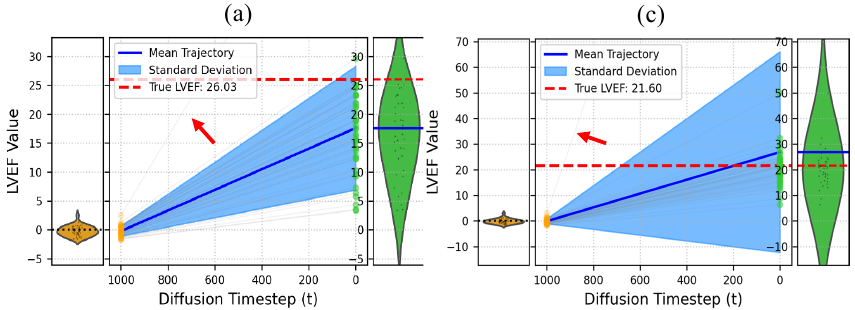}
    \caption{\textbf{Failure Mode Visualization.} In high-error cases, the conditional guidance is often insufficient to unify the paths. Red arrows highlight divergent trajectories that break away from the main cluster, serving as a visual indicator of low reliability.}
    \label{fig:failure_modes}
\end{figure}

\textbf{Visualizing Failure Modes.}
The red arrows in Figure~\ref{fig:failure_modes} highlight a phenomenon of conditional guidance failure. In these challenging cases, the learned gradient field is too weak to uniformly guide all initialization points towards the central mode. Consequently, a subset of trajectories—originating from specific regions of the Gaussian prior—diverges significantly from the main cluster. This divergence suggests that the conditional probability landscape is flat or multi-modal, meaning the evidence supports multiple potential interpretations.

\section{Discussion}
\label{sec:discussion}

In this work, we proposed a paradigm shift from deterministic to generative regression for LVEF estimation. While the experimental results demonstrate state-of-the-art performance, several aspects regarding clinical practicality, model behavior, and limitations warrant deeper discussion.

\subsection{Computational Cost vs. Clinical Safety}
A valid concern arises regarding the inference latency introduced by the generative process. As noted in Table~5 (in Experiments), while a single DDIM step is extremely fast (0.02s), estimating a reliable posterior distribution requires sampling $K$ trajectories (e.g., $K=40$). A sequential execution would indeed imply a total latency of roughly 0.8s per patient, which is higher than the 0.01s of a deterministic baseline.

We argue that this computational trade-off is both justifiable and negligible in clinical practice for two primary reasons. First, the $K$ stochastic trajectories are strictly independent given the condition $\mathbf{c}$, allowing them to be processed as a single batch operation on modern GPUs. Because our MCSN architecture is exceptionally lightweight, with only 0.69M parameters, increasing the batch size from 1 to 40 incurs minimal overhead; thus, the total wall-clock time remains well within real-time requirements at less than 0.1s. Second, in high-stakes medical decision-making, the cost of a misdiagnosis far outweighs any marginal increase in computing time. While a deterministic model might execute in 0.01s, its inability to signal failure on noisy inputs poses a severe safety risk. Conversely, our framework operates in sub-second time while explicitly flagging unreliable predictions through high variance. We posit that this safety buffer is an essential requirement for trustworthy AI deployment, validating the modest increase in computational demand.

\subsection{Physiological Boundaries in Gaussian Diffusion}
Our current diffusion framework operates in an unbounded Gaussian space. Consequently, there is no inherent mathematical constraint preventing the generated scalar $\hat{y}$ from falling outside the physiological range of LVEF (i.e., $[0, 100]$). In our experiments, we observed that for atypical hearts, the model occasionally predicts values slightly exceeding 100\%.

This phenomenon stems from the tail behavior of the Gaussian prior. While one could impose bounded activations (e.g., Sigmoid) or truncated diffusion schemes, these often introduce gradient vanishing issues or complicate the score matching objective. In practice, we find that a simple post-hoc operation---clipping the final mean prediction to $[0, 100]$---is sufficient and does not degrade performance. Importantly, the occurrence of out-of-bound samples is often informative in itself: it typically correlates with high aleatoric uncertainty, suggesting that the model is struggling to locate the exact manifold of the data due to extreme input features.

% --- Figure 9: Failure Modes ---

\subsection{Limitations and Future Directions}
Despite the promising results, our study has limitations that outline directions for future research.

\textbf{Dependency on Complete Modalities.} Our dual-encoder architecture currently assumes the availability of both visual data and tabular attributes. In clinical practice, demographic records may occasionally be incomplete (the missing modality problem). Currently, our model processes attributes as a holistic vector. Future work will explore modality-dropout training strategies to enhance robustness against missing clinical priors.

\textbf{2D vs. 3D Assessment.} Like most deep learning approaches in this domain, MCSDR estimates LVEF from 2D video slices (A4C view). While this is the clinical standard for initial assessment, it relies on geometric assumptions about the left ventricle. Extending our conditional score-based framework to 3D echocardiography could theoretically provide a more direct volumetric assessment, provided that the scarcity of labeled 3D datasets can be addressed.

\section{Conclusion}
\label{sec:conclusion}

This paper presents the Multimodal Conditional Score-based Diffusion model for Regression (MCSDR), a novel framework that reformulates LVEF estimation as a probabilistic generative task. Recognizing that extracting scalar metrics from noisy ultrasound images is inherently an ill-posed inverse problem, we move beyond the limitations of deterministic point estimation to model the full conditional posterior distribution of the target variable.
Our methodology introduces the Multimodal Conditional Score Network (MCSN), which is specifically designed to integrate high-dimensional spatiotemporal visual representations with tabular clinical priors. This architecture allows the model to leverage demographic context to constrain the solution space, guiding the diffusion-based sampling process toward a robust recovery of ground-truth LVEF values. Extensive experiments conducted on three diverse datasets—EchoNet-Dynamic, EchoNet-Pediatric, and CAMUS—demonstrate that MCSDR achieves new state-of-the-art performance. 
Furthermore, by visualizing generative trajectories and posterior landscapes, our framework provides a novel interpretable perspective for reliable physiological signal assessment.

\bibliographystyle{ieeetr}
\bibliography{main}

@article{circulation_1971_lvef,
author = {JOAQUIN F. POMBO  and BART L. TROY  and RICHARD O. RUSSELL },
title = {Left Ventricular Volumes and Ejection Fraction by Echocardiography},
journal = {Circulation},
volume = {43},
number = {4},
pages = {480-490},
year = {1971},
doi = {10.1161/01.CIR.43.4.480},
URL = {https://www.ahajournals.org/doi/abs/10.1161/01.CIR.43.4.480},
eprint = {https://www.ahajournals.org/doi/pdf/10.1161/01.CIR.43.4.480}}

@article{jacc_2003_lvef,
author = {Jeptha P Curtis  and Seth I Sokol  and Yongfei Wang  and Saif S Rathore  and Dennis T Ko  and Farid Jadbabaie  and Edward L Portnay  and Stephen J Marshalko  and Martha J Radford  and Harlan M Krumholz },
title = {The association of left ventricular ejection fraction, mortality, and cause of death in stable outpatients with heart failure},
journal = {JACC},
volume = {42},
number = {4},
pages = {736-742},
year = {2003},
doi = {10.1016/S0735-1097(03)00789-7},

URL = {https://www.jacc.org/doi/abs/10.1016/S0735-1097%2803%2900789-7},
eprint = {https://www.jacc.org/doi/pdf/10.1016/S0735-1097%2803%2900789-7}}

@article{jacc_2009_lvef,
author = {Micha T. Maeder  and David M. Kaye },
title = {Heart Failure With Normal Left Ventricular Ejection Fraction},
journal = {JACC},
volume = {53},
number = {11},
pages = {905-918},
year = {2009},
doi = {10.1016/j.jacc.2008.12.007},

URL = {https://www.jacc.org/doi/abs/10.1016/j.jacc.2008.12.007},
eprint = {https://www.jacc.org/doi/pdf/10.1016/j.jacc.2008.12.007}

}

@article{EchoNet-Dynamic,
  title={Video-based AI for beat-to-beat assessment of cardiac function},
  author={Ouyang, David and He, Bryan and Ghorbani, Amirata and Yuan, Neal and Ebinger, Joseph and Langlotz, Curtis P and Heidenreich, Paul A and Harrington, Robert A and Liang, David H and Ashley, Euan A and others},
  journal={Nature},
  volume={580},
  number={7802},
  pages={252--256},
  year={2020},
  publisher={Nature Publishing Group UK London}
}

@article{EchoNet-Pediatric,
  title={Video-based deep learning for automated assessment of left ventricular ejection fraction in pediatric patients},
  author={Reddy, Charitha D and Lopez, Leo and Ouyang, David and Zou, James Y and He, Bryan},
  journal={Journal of the American Society of Echocardiography},
  volume={36},
  number={5},
  pages={482--489},
  year={2023},
  publisher={Elsevier}
}

@book{bishop2006pattern,
  title={Pattern recognition and machine learning},
  author={Bishop, Christopher M and Nasrabadi, Nasser M},
  volume={4},
  number={4},
  year={2006},
  publisher={Springer}
}

@ARTICLE{CAMUS,
  author={Leclerc, Sarah and Smistad, Erik and Pedrosa, João and Østvik, Andreas and Cervenansky, Frederic and Espinosa, Florian and Espeland, Torvald and Berg, Erik Andreas Rye and Jodoin, Pierre-Marc and Grenier, Thomas and Lartizien, Carole and D’hooge, Jan and Lovstakken, Lasse and Bernard, Olivier},
  journal={IEEE Transactions on Medical Imaging}, 
  title={Deep Learning for Segmentation Using an Open Large-Scale Dataset in 2D Echocardiography}, 
  year={2019},
  volume={38},
  number={9},
  pages={2198-2210},
  doi={10.1109/TMI.2019.2900516}}

@article{circulation_2022_lvef,
author = {Sebastian Rosch  and Karl-Patrik Kresoja  and Christian Besler  and Karl Fengler  and Anne Rebecca Schöber  and Maximilian von Roeder  and Christian Lücke  and Matthias Gutberlet  and Karin Klingel  and Holger Thiele  and Karl-Philipp Rommel  and Philipp Lurz },
title = {Characteristics of Heart Failure With Preserved Ejection Fraction Across the Range of Left Ventricular Ejection Fraction},
journal = {Circulation},
volume = {146},
number = {7},
pages = {506-518},
year = {2022},
doi = {10.1161/CIRCULATIONAHA.122.059280},
URL = {https://www.ahajournals.org/doi/abs/10.1161/CIRCULATIONAHA.122.059280},
eprint = {https://www.ahajournals.org/doi/pdf/10.1161/CIRCULATIONAHA.122.059280}}

@article{EchoClip,
  title={Vision--language foundation model for echocardiogram interpretation},
  author={Christensen, Matthew and Vukadinovic, Milos and Yuan, Neal and Ouyang, David},
  journal={Nature Medicine},
  pages={1--8},
  year={2024},
  publisher={Nature Publishing Group US New York}
}

@ARTICLE{tip2013,
  author={Alessandrini, Martino and Basarab, Adrian and Liebgott, Hervé and Bernard, Olivier},
  journal={IEEE Transactions on Image Processing}, 
  title={Myocardial Motion Estimation From Medical Images Using the Monogenic Signal}, 
  year={2013},
  volume={22},
  number={3},
  pages={1084-1095},
  doi={10.1109/TIP.2012.2226903}}

@ARTICLE{tip2005,
  author={Suhling, M. and Arigovindan, M. and Jansen, C. and Hunziker, P. and Unser, M.},
  journal={IEEE Transactions on Image Processing}, 
  title={Myocardial motion analysis from B-mode echocardiograms}, 
  year={2005},
  volume={14},
  number={4},
  pages={525-536},
  doi={10.1109/TIP.2004.838709}}

@ARTICLE{tip2023,
  author={Ouzir, Nora and Basarab, Adrian and Liebgott, Hervé and Harbaoui, Brahim and Tourneret, Jean-Yves},
  journal={IEEE Transactions on Image Processing}, 
  title={Motion Estimation in Echocardiography Using Sparse Representation and Dictionary Learning}, 
  year={2018},
  volume={27},
  number={1},
  pages={64-77},
  doi={10.1109/TIP.2017.2753406}}

@INPROCEEDINGS{RU-Unet,
  author={Leclerc, Sarah and Smistad, Erik and Grenier, Thomas and Lartizien, Carole and Ostvik, Andreas and Cervenansky, Frederic and Espinosa, Florian and Espeland, Torvald and Rye Berg, Erik Andreas and Jodoin, Pierre-Marc and Lovstakken, Lasse and Bernard, Olivier},
  booktitle={2019 IEEE International Ultrasonics Symposium (IUS)}, 
  title={RU-Net: A refining segmentation network for 2D echocardiography}, 
  year={2019},
  volume={},
  number={},
  pages={1160-1163},
  doi={10.1109/ULTSYM.2019.8926158}}

@article{simlvseg,
title = {SimLVSeg: Simplifying Left Ventricular Segmentation in 2-D+Time Echocardiograms With Self- and Weakly Supervised Learning},
journal = {Ultrasound in Medicine \& Biology},
volume = {50},
number = {12},
pages = {1945-1954},
year = {2024},
issn = {0301-5629},
doi = {https://doi.org/10.1016/j.ultrasmedbio.2024.08.023},
url = {https://www.sciencedirect.com/science/article/pii/S030156292400334X},
author = {Fadillah Maani and Asim Ukaye and Nada Saadi and Numan Saeed and Mohammad Yaqub},
}

@InProceedings{miccai_2021_UVT,
author="Reynaud, Hadrien
and Vlontzos, Athanasios
and Hou, Benjamin
and Beqiri, Arian
and Leeson, Paul
and Kainz, Bernhard",
editor="de Bruijne, Marleen
and Cattin, Philippe C.
and Cotin, St{\'e}phane
and Padoy, Nicolas
and Speidel, Stefanie
and Zheng, Yefeng
and Essert, Caroline",
title="Ultrasound Video Transformers for Cardiac Ejection Fraction Estimation",
booktitle="Medical Image Computing and Computer Assisted Intervention -- MICCAI 2021",
year="2021",
publisher="Springer International Publishing",
address="Cham",
pages="495--505",
isbn="978-3-030-87231-1"
}

@InProceedings{Miccai_2022_echognn,
author="Mokhtari, Masoud
and Tsang, Teresa
and Abolmaesumi, Purang
and Liao, Renjie",
editor="Wang, Linwei
and Dou, Qi
and Fletcher, P. Thomas
and Speidel, Stefanie
and Li, Shuo",
title="EchoGNN: Explainable Ejection Fraction Estimation with Graph Neural Networks",
booktitle="Medical Image Computing and Computer Assisted Intervention -- MICCAI 2022",
year="2022",
publisher="Springer Nature Switzerland",
address="Cham",
pages="360--369",
isbn="978-3-031-16440-8"
}

@InProceedings{Miccai_2022_echograph,
author="Thomas, Sarina
and Gilbert, Andrew
and Ben-Yosef, Guy",
editor="Wang, Linwei
and Dou, Qi
and Fletcher, P. Thomas
and Speidel, Stefanie
and Li, Shuo",
title="Light-weight Spatio-Temporal Graphs for Segmentation and Ejection Fraction Prediction in Cardiac Ultrasound",
booktitle="Medical Image Computing and Computer Assisted Intervention -- MICCAI 2022",
year="2022",
publisher="Springer Nature Switzerland",
address="Cham",
pages="380--390",
isbn="978-3-031-16440-8"
}

@InProceedings{miccai_2022_echocotr,
	author="Muhtaseb, Rand and Yaqub, Mohammad",
	editor="Wang, Linwei and Dou, Qi and Fletcher, P. Thomas and Speidel, Stefanie and Li, Shuo",
	title="EchoCoTr: Estimation of the Left Ventricular Ejection Fraction from Spatiotemporal Echocardiography",
	booktitle="Medical Image Computing and Computer Assisted Intervention -- MICCAI 2022",
	year="2022",
	publisher="Springer Nature Switzerland",
	address="Cham",
	pages="370--379",
isbn="978-3-031-16440-8"
}

@Article{diag_2023_Batool,
AUTHOR = {Batool, Samana and Taj, Imtiaz Ahmad and Ghafoor, Mubeen},
TITLE = {Ejection Fraction Estimation from Echocardiograms Using Optimal Left Ventricle Feature Extraction Based on Clinical Methods},
JOURNAL = {Diagnostics},
VOLUME = {13},
YEAR = {2023},
NUMBER = {13},
ARTICLE-NUMBER = {2155},
URL = {https://www.mdpi.com/2075-4418/13/13/2155},
PubMedID = {37443550},
ISSN = {2075-4418},
DOI = {10.3390/diagnostics13132155}
}

@InProceedings{miccai_coreecho,
author="Maani, Fadillah Adamsyah
and Saeed, Numan
and Matsun, Aleksandr
and Yaqub, Mohammad",
editor="Linguraru, Marius George
and Dou, Qi
and Feragen, Aasa
and Giannarou, Stamatia
and Glocker, Ben
and Lekadir, Karim
and Schnabel, Julia A.",
title="CoReEcho: Continuous Representation Learning for 2D+Time Echocardiography Analysis",
booktitle="Medical Image Computing and Computer Assisted Intervention -- MICCAI 2024",
year="2024",
publisher="Springer Nature Switzerland",
address="Cham",
pages="591--601",
isbn="978-3-031-72083-3"
}

@ARTICLE{tmi_2023_semilvef,
  author={Dai, Weihang and Li, Xiaomeng and Ding, Xinpeng and Cheng, Kwang-Ting},
  journal={IEEE Transactions on Medical Imaging}, 
  title={Cyclical Self-Supervision for Semi-Supervised Ejection Fraction Prediction From Echocardiogram Videos}, 
  year={2023},
  volume={42},
  number={5},
  pages={1446-1461},
  doi={10.1109/TMI.2022.3229136}}

@ARTICLE{tim_2025_fhfa,
  author={Mo, Haimiao and Liang, Juan and Li, Bing and Hu, Zhijian and Yi, Meng and Wu, Hongjia and Rong, Qian and Xu, Zeyuan},
  journal={IEEE Transactions on Instrumentation and Measurement}, 
  title={A Noninvasive Framework for Heart Function Assessment by Multitask Learning}, 
  year={2025},
  volume={74},
  number={},
  pages={1-13},
  doi={10.1109/TIM.2025.3553885}}

@misc{icml_2025_tutorial,
  title     = {Harnessing Low Dimensionality in Diffusion Models: From Theory to Practice},
  author    = {Yuxin Chen and Qing Qu and Liyue Shen},
  year      = {2025},
  note      = {International Conference on Learning Representations},
  url       = {https://icml-2025-tutorial.github.io/diffusion-models/}, 
}

@article{song2019generative,
  title={Generative modeling by estimating gradients of the data distribution},
  author={Song, Yang and Ermon, Stefano},
  journal={Advances in Neural Information Processing Systems},
  volume={32},
  year={2019}
}

@article{hyvarinen2005estimation,
  title={Estimation of non-normalized statistical models by score matching},
  author={Hyv{\"a}rinen, Aapo and Dayan, Peter},
  journal={Journal of Machine Learning Research},
  volume={6},
  number={4},
  year={2005}
}

@article{vincent2011connection,
  title={A connection between score matching and denoising autoencoders},
  author={Vincent, Pascal},
  journal={Neural Computation},
  volume={23},
  number={7},
  pages={1661--1674},
  year={2011},
  publisher={MIT Press}
}

@inproceedings{song2020sliced,
  title={Sliced score matching: A scalable approach to density and score estimation},
  author={Song, Yang and Garg, Sahaj and Shi, Jiaxin and Ermon, Stefano},
  booktitle={Proceedings of the 36th Conference on Uncertainty in Artificial Intelligence (UAI)},
  pages={574--584},
  year={2020},
  organization={PMLR}
}

@inproceedings{song2021scorebased,
  title={Score-Based Generative Modeling through Stochastic Differential Equations},
  author={Song, Yang and Sohl-Dickstein, Jascha and Kingma, Diederik P and Kumar, Abhishek and Ermon, Stefano and Poole, Ben},
  booktitle={International Conference on Learning Representations (ICLR)},
  year={2021}
}

@inproceedings{ho2020denoising,
  title={Denoising diffusion probabilistic models},
  author={Ho, Jonathan and Jain, Ajay and Abbeel, Pieter},
  booktitle={Advances in Neural Information Processing Systems},
  volume={33},
  pages={6840--6851},
  year={2020}
}

@inproceedings{DDIM2021,
  author    = {iaming Song, Chenlin Meng and Stefano Ermon},
  title     = {DENOISING DIFFUSION IMPLICIT MODELS},
  booktitle = {International Conference on Learning Representations (ICLR)},
  year      = {2021},
}

@inproceedings{song2020improved,
  title={Improved techniques for training score-based generative models},
  author={Song, Yang and Ermon, Stefano},
  booktitle={Advances in Neural Information Processing Systems},
  volume={33},
  pages={12438--12448},
  year={2020}
}

@inproceedings{lu2022dpm,
  title={DPM-Solver: A fast ODE solver for diffusion probabilistic model sampling in around 10 steps},
  author={Lu, Cheng and Zhou, Yuhao and Bao, Fan and Chen, Jianfei and Li, Chongxuan and Zhu, Jun},
  booktitle={Advances in Neural Information Processing Systems},
  volume={35},
  pages={5775--5787},
  year={2022}
}

@article{lu2025dpm,
  title={DPM-Solver++: Fast solver for guided sampling of diffusion probabilistic models},
  author={Lu, Cheng and Zhou, Yuhao and Bao, Fan and Chen, Jianfei and Li, Chongxuan and Zhu, Jun},
  journal={Machine Intelligence Research},
  pages={1--22},
  year={2025},
  publisher={Springer}
}

@inproceedings{song2021maximum,
  title={Maximum likelihood training of score-based diffusion models},
  author={Song, Yang and Durkan, Conor and Murray, Iain and Ermon, Stefano},
  booktitle={Advances in Neural Information Processing Systems},
  volume={34},
  pages={1415--1428},
  year={2021}
}

@article{efron2011tweedie,
  title={Tweedie’s formula and selection bias},
  author={Efron, Bradley},
  journal={Journal of the American Statistical Association},
  volume={106},
  number={496},
  pages={1602--1614},
  year={2011},
  publisher={Taylor \& Francis}
}

@inproceedings{sohl2015deep,
  title={Deep unsupervised learning using nonequilibrium thermodynamics},
  author={Sohl-Dickstein, Jascha and Weiss, Eric and Maheswaranathan, Niru and Ganguli, Surya},
  booktitle={International Conference on Machine Learning (ICML)},
  pages={2256--2265},
  year={2015},
  organization={PMLR}
}

@inproceedings{tashiro2021csdi,
  title={CSDI: Conditional score-based diffusion models for probabilistic time series imputation},
  author={Tashiro, Yusuke and Song, Jiaming and Song, Yang and Ermon, Stefano},
  booktitle={Advances in Neural Information Processing Systems},
  volume={34},
  pages={24804--24816},
  year={2021}
}

@inproceedings{zhang2023adding,
  title={Adding conditional control to text-to-image diffusion models},
  author={Zhang, Lvmin and Rao, Anyi and Agrawala, Maneesh},
  booktitle={Proceedings of the IEEE/CVF International Conference on Computer Vision (ICCV)},
  pages={3836--3847},
  year={2023}
}

@article{panecho,
    author = {Holste, Gregory and Oikonomou, Evangelos K. and Tokodi, Márton and Kovács, Attila and Wang, Zhangyang and Khera, Rohan},
    title = {Complete AI-Enabled Echocardiography Interpretation With Multitask Deep Learning},
    journal = {JAMA},
    volume = {334},
    number = {4},
    pages = {306-318},
    year = {2025},
    month = {07},
    issn = {0098-7484},
    doi = {10.1001/jama.2025.8731},
    url = {https://doi.org/10.1001/jama.2025.8731},
    eprint = {https://jamanetwork.com/journals/jama/articlepdf/2835630/jama_holste_2025_oi_250039_1753124028.63327.pdf},
}

% \newpage

 

% \end{IEEEbiography}



% \vfill

\end{document}